Research article

# Multimodal fused deep learning for drug property prediction: Integrating chemical language and molecular graph

Xiaohua Lu, Liangxu Xie *, Lei Xu, Rongzhi Mao, Xiaojun Xu *, Shan Chang *

*Institute of Bioinformatics and Medical Engineering, Jiangsu University of Technology, Changzhou 213001, China*

A R T I C L E I N F O

A B S T R A C T

Accurately predicting molecular properties is a challenging but essential task in drug discovery. Recently, many mono-modal deep learning methods have been successfully applied to molecular property prediction. However, mono-modal learning is inherently limited as it relies solely on a single modality of molecular representation, which restricts a comprehensive understanding of drug molecules. To overcome the limitations, we propose a multimodal fused deep learning (MMFDL) model to leverage information from different molecular representations. Specifically, we construct a triple-modal learning model by employing Transformer-Encoder, Bidirectional Gated Recurrent Unit (BiGRU), and graph convolutional network (GCN) to process three modalities of information from chemical language and molecular graph: SMILES-encoded vectors, ECFP fingerprints, and molecular graphs, respectively. We evaluate the proposed triple-modal model using five fusion approaches on six molecule datasets, including Delaney, Llinas2020, Lipophilicity, SAMPL, BACE, and pKa from DataWarrior. The results show that the MMFDL model achieves the highest Pearson coefficients, and stable distribution of Pearson coefficients in the random splitting test, outperforming mono-modal models in accuracy and reliability. Furthermore, we validate the generalization ability of our model in the prediction of binding constants for protein-ligand complex molecules, and assess the resilience capability against noise. Through analysis of feature distributions in chemical space and the assigned contribution of each modal model, we demonstrate that the MMFDL model shows the ability to acquire complementary information by using proper models and suitable fusion approaches. By leveraging diverse sources of bioinformatics information, multimodal deep learning models hold the potential for successful drug discovery.

## 1. Introduction

Drug discovery has historically been costly with low success rates [1]. Many factors need to be considered for successful drug discovery, including solubility [2], lipophilicity [3], inhibitory activity [4], acidity/basicity [5], and assessing protein-ligand binding affinity [6]. Experimental determination is often expensive and time-consuming. Accurate prediction of drug properties is a crucial aspect of drug discovery [7]. Computational methods have been established to assist drug discovery, including physics-based and machine learning-based methods [8,9]. Physics-based computational methods, including molecular docking [10], molecular dynamics simulations [11], and quantitative structure-activity relationships (QSAR) [12,13], excel in examining molecular structure, interaction, and motion, making significant contributions to drug research and development. However, physics-based computational methods face challenges, such as computational complexity, high resource requirements, and reliance on experimental data [14]. With the improvement of computer performance and calculation methods, deep learning has emerged as a promising approach, achieving notable success in predicting drug properties [15–20].

To apply deep learning to drug property prediction, molecules need to be transformed into molecular representations [21,22]. Researchers have explored various representation methods, including Simplified Molecular Input Line Entry System (SMILES), extended connectivity fingerprints (ECFP), molecular graphs, etc. The application of molecular representations in machine / deep learning models has effectively improved QSAR prediction performance [23]. For example, Karimi et al. constructed the DeepAffinity model using integer encoding for SMILES and numeric encoding for amino acid sequences [24]. ECFP is one type of topological-based method that converts the neighborhood information into bit strings. ECFP and its variants have been widely used in bioinformatics and drug discovery [25,26]. SMILES-encoded vectors and ECFP are numerical representations used to encode chemical structures [27,28]. They can be considered as the chemical language [29]. To capture the whole topology of molecules, another type of molecular representation called molecular graph is proposed [30]. In molecular graphs, atoms are described as nodes, and bonds are represented as edges [31]. Various graph neural networks (GNN) have been employed to process molecular graphs [30], including graph neural networks[32], message-passing neural network (MPNN) [33], and variants with attention mechanisms [34,35] to predict molecular properties. Different molecular representations provide particular types of information. Interdisciplinary work is required to obtain the specific information conveyed from molecular representations.

* Corresponding authors.
*E-mail addresses:* xieliangxu@jsut.edu.cn (L. Xie), xuxiaojun@jsut.edu.cn (X. Xu), schang@jsut.edu.cn (S. Chang).

Though previously reported methods have made progress using single-input modes or mono-modal models, mono-modal learning suffers from incomplete information representation. Mono-modal learning exhibits dependence on dataset and model selection. To enhance the expressiveness of existing molecular representations and deep learning models, combined molecular representation learning methods, multi-task learning, and ensemble methods have been developed [13, 36–41]. Currently, multi-modal learning is a vibrant multidisciplinary field, which provides frameworks to process multiple sources of information [42,43]. Multimodal learning is a general approach for incorporating artificial intelligence models that can extract and associate information from multimodal data, enabling models to handle complex relationships between different modalities [44,45]. The rich and diverse information in multimodal data is essential for drug property prediction [46]. Several pioneering works have been conducted recently. For example, Dong et al. designed the multimodal attribute learning framework (MMA-DPI) using the molecular transformer and graph convolutional networks for predicting drug-protein interactions, and experimental results showed that the proposed method achieved higher performance than the existing advanced frameworks [47]. Sun et al. proposed bi-modal representations for protein-protein interactions, which achieved higher screening ability [48]. The great potential of multimodal descriptors for prediction of pharmacokinetic properties has been demonstrated. For example, Iwata et al. proposed multimodal learning methods to predict drug clearance by using molecular structure-based descriptors (imputed descriptors from chemical structure, and graph data), as well as animal data, and nonclinical data, respectively [49,50]. Handa et al. demonstrated multimodal learning in prediction Tissue-to-plasma partition coefficient using physicochemical descriptors and minimum required experimental values [51]. Their findings indicate that multimodal models offer improved prediction accuracy and can be applied to various scenarios, suggesting the advantage of multimodal models in integrating different sources of data for improved predictive performance.

When processing more than one molecular descriptors, aggregation function is required. For example, a popular machine learning package, Chemprop, provides three options to aggregate molecular features: summation, a scaled sum (divided by a specified scaler) and average [52]. Besides using the mentioned aggregation function, multi-modal fusion method provides more options to effectively integrate information from different data sources. Multimodal approaches have been employed in the field of drug discovery. However, it is essential to adopt proper approach to leverage or balance contribution from different modalities because the information can exist as redundant, complementary, or cooperative [53]. Existing multimodal techniques include early and late fusion [54], hybrid fusion, model ensemble [55], and more recently, joint training methods based on deep learning networks [56,57]. Multimodal systems need to ingest, interpret, and reason from multiple modal information sources. Therefore, multimodal methods pose challenges, including the representation of multimodal data, modality transformations, integrating modality information, measuring relationships between different modalities, and transferring knowledge between modality information and their predictive models. Several attempts have been made to address the challenges [43]. To filter noise while capturing synergies from different modalities, Liu et al. proposed a novel penalty-modality method, making decisions independently for each modality first, and then completing a multimodal combination in a differentiable or multiplicative manner, which effectively improves the accuracy of the prediction [58]. Yang et al. also emphasize the importance of the fusion strategy [59]. Dehghan et al. propose a triplet loss function in the processing of multimodal representation learning [60].

Designing a multimodal deep learning method suitable for drug property prediction holds significant promise to enhance the accuracy of existing computational approaches and broaden their applicability, presenting important prospects for practical applications. In this study, we propose a triple-modal fusion learning method to predict molecular properties. Two major contributions of this study are summarized as follows:

- We employed Transformer-Encoder, BiGRU, and GCN models to process SMILES-encoded vectors, ECFP, and molecular graphs. In bioinformatics, molecules are generally represented by these three different and complementary representations. The proposed multimodal fused deep learning model can harness diverse information while fully utilizing the advantage of the deep learning models.
- To leverage information from multimodal features, proper fusion approaches are required. We select the fusion approaches upon the previous study [58], and adopt and evaluate four machine learning methods (LASSO, Elastic Net, Gradient boosting (GB), and random forest (RF)), along with one numerical method called stochastic gradient descent (SGD). These methods are employed to assign the contributions for each modal learning.

We compared the performance of the proposed multimodal fusion deep learning (MMFDL) model with mono-modal models in predicting drug properties. Experimental results indicate that our constructed MMFDL improves predictive performance, as well as increasing generalization ability and enhancing noise resistance.

## 2. Materials and methods

### 2.1. Datasets

We evaluate the proposed multimodal modal model on six single-molecule drug property datasets and one protein-ligand complex molecule dataset, including Delaney, Llinas2020, Lipophilicity, SAMPL, BACE, pKa from DataWarrior, and the refined set of PDBbind [61,62]. The relevant properties of datasets can be found in Table S1 in Supporting Information (SI).

(a) Delaney is a dataset for molecular solubility, comprising the chemical structures and corresponding experimentally measured solubility data for 1128 small molecules [61]. The solubility values in the data set are log-transformed values.
(b) Llinas2020 dataset is derived from the SolTransNet dataset [63]. The training set contains 9860 molecules, and the test set of Llinas2020 has 100 molecules with solubility values according to the reported splitting way [63]. The solubility values in the data set are log-transformed values.
(c) Lipophilicity is a lipophilic property dataset from MoleculeNet, consisting of logD at pH 7.4 and SMILES for 4200 molecules [61].
(d) SAMPL dataset from MoleculeNet is a statistical evaluation dataset for small molecule modeling and consists of 642 experimental values logP [61].
(e) BACE is a dataset on the activity of compounds against β-secretase, and contains data on the activity of 1513 compounds against β-secretase [61].
(f) pKa data provided by the DataWarrior visualization software proposed by Sander et al. contains 7910 data related to pKa [62].

(g) The refined set of PDBbind v2020 dataset after data cleaning and molecular representation conversion has a total of 5089 binding constants for biological molecular complexes [61].

Except for the Llinas2020 dataset that is divided according to previous division [64], other datasets were initially divided into training and test sets in a 9:1 ratio. Additionally, the tuning set is randomly selected from the training set, with a ratio of 2:8. The data in the training set is used for learning, and the tuning set is used for assigning weights for the contribution from triple-modal features, while the data in the test set consists of molecules unseen by the deep neural network, serving to evaluate the learning capabilities of the trained model. To check the reliability of the model during the repeated training process, we employ different random seeds to split the dataset into training and test datasets.

## 2.2. Multimodal input

Previous studies by Handa et al. have highlighted the significant contribution of chemical structure to the prediction of pharmacokinetic properties [51]. This suggests that chemical structure-based descriptors play a crucial role not only in predicting pharmacokinetic properties but also in forecasting other drug-related characteristics. We utilize three different representations to represent multimodal inputs, as illustrated in Fig. 1, which are SMILES-encoded vectors, ECFP fingerprints, and 2D molecular graphs. SMILES-encoded vectors can obtain the basic topological and chemical information of molecules. ECFP is a universal molecular fingerprint, which can provide substructure topological information of molecules. The molecular graph can comprehensively describe the relationship between atoms, spatial distribution, and other topological information. SMILES-encoded vectors and ECFP use sequential numerical values to describe molecules, which can be considered as chemical language.

In bioinformatics and cheminformatics, SMILES, ECFP, and molecular graphs, are the most commonly used to represent molecules. SMILES is a textual representation of chemical structures where each character represents a specific atom or bond in the molecule. ECFP fingerprints capture local chemical features and spatial information. 2D molecular graphs provide details on atom types, connectivity, and chemical bonding. Recognizing the differences among these representations, we propose constructing a multimodal model integrating SMILES, ECFP, and molecular graphs to leverage their complementary information. We select appropriate models for each representation: Transformer models for SMILES-encoded vectors due to their effectiveness in processing sequential data and capturing inter-atomic spatial relationships, and BiGRU for ECFP representations owing to their ability to capture hierarchical and sequential relationships encoded by ECFP, and Graph Convolutional Networks (GCN) for molecular graphs, leveraging their effectiveness in capturing node connections and feature information. Through this approach, we aim to enhance molecular representation and improve predictive performance in bioinformatics and cheminformatics applications.

### 2.2.1. SMILES tokenization and vectorization

SMILES is a line symbol that represents the atoms, bonds, and rings that make up a molecule as a string. The detailed specification of SMILES can be found in OpenSMILES [65]. SMILES tokenization and vectorization are revised based on custom regular expressions that can capture the 'single atom' type, 'double bond' type, 'triple bond' type, etc. For example, the SMILES string 'CS(=O)(=O)Cl' uses regular expressions and can be split into a list of ['C', 'S', '(', '=', 'O', ')', '(', '=', 'O', ')', 'Cl']. Converting molecular structural information into numerical form facilitates a variety of tasks in deep learning [27].

For different datasets, a unified regular expression is used to construct a label dictionary. The SMILES in the datasets are mapped to fixed-length integer sequences by the labeling dictionary (refer to Fig. S1 in SI for the length distribution of SMILES strings). Every token in the SMILES sequence is subjected to encoding, followed by the generation of an embedding vector for each token. Any shorter SMILES strings were padded with zeros at the end. The length of SMILES-encoded vectors in different data is shown in Table S2.

### 2.2.2. ECFP Fingerprints

ECFP is a circular topological fingerprint that uses circular atomic neighborhoods to generate variable-length hashed integer identifiers, with each unique identifier corresponding to a unique substructure of the compound [28]. This approach describes a molecule as a fixed-length bit string, with each bit indicating the presence or absence of a particular substructure in the molecule. This fingerprinting method [28] can be used for structure-activity modeling, Quantitative Structure-Activity Relationship (QSAR) analysis, and drug ADME/T prediction. Several software programs providing fingerprints include Pipe-line Pilot [66], Chemaxon's JChem [67], CDK [68], and RDKit [69]. The ECFP fingerprints that we used has a length of 1024 with the radius of 2. The lengths of ECFP fingerprints in different datasets are shown in Table S2.

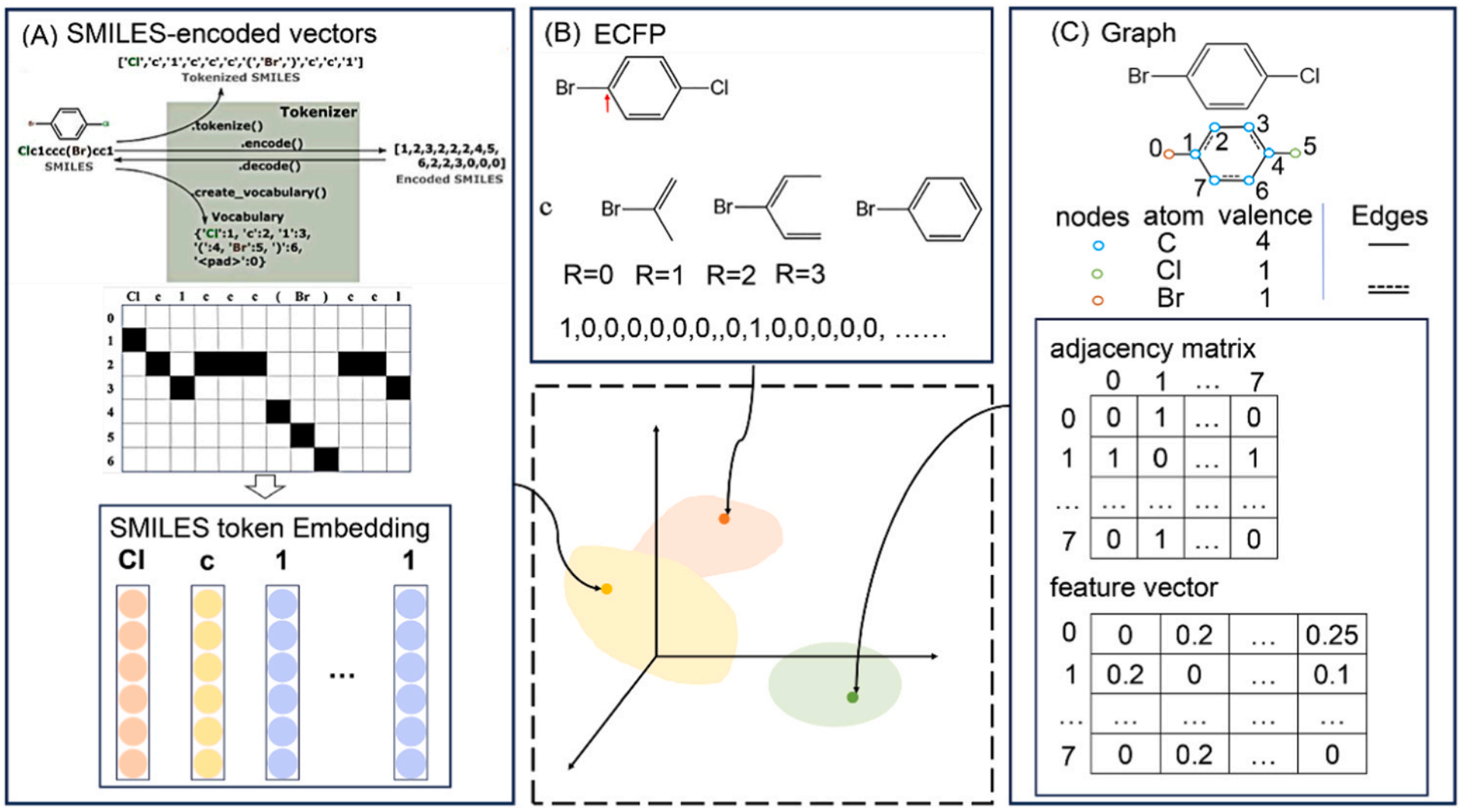

Fig. 1. Molecular representations of molecules. Molecules are described as SMILES-encoded vectors, ECFP, and molecular graphs. The schematic figure shows that three molecular representations may be distributed in different regions of chemical spaces.

### 2.2.3. Graph

We represent molecules as graphs to obtain more accurate structural topology information about the molecules [31]. A molecular graph can be formulated as $G = (V, E)$, where $V$ is a set of n atomic nodes, with each node represented by a feature vector composed of 78 values, including 44 atom types, 11 one-hot encoding atom degree, 11 one-hot encoding of the total number of hydrogens, 11 one-hot encoding valences and 1 bit for the aromaticity. $E$ is the set of edges represented by the adjacency matrix of the molecular graph. The existence of an edge in an adjacency matrix depends on whether there are direct covalent chemical bonds between the corresponding atoms in the molecule.

## 2.3. Multimodal learning model

The overall framework of our proposed multimodal fused deep learning (MMFDL) models is illustrated in Fig. 2. The source code is available at https://github.com/AIMedDrug/MMFDL.git. As depicted in Fig. 2A, for learning SMILES-encoded vectors, we use two identical primitive Transformer-Encoders, and the output is sent to the fully connected layer. As shown in Fig. 2B, for learning ECFP fingerprints, we send the ECFP through two layers of BiGRU and multi-head attention layers to the fully connected layer, which we comprehensively represent this model as BiGRU. We utilize three similar representations to process multi-modal inputs: SMILES-encoded vectors, ECFP fingerprints, and molecular graphs. Three representations are similar, but can offer different information. Future work can incorporate other types of molecular descriptors if needed. As shown in Fig. 2C, for learning molecular graph information, we use the graph's adjacency matrix and feature vectors as input, employ two GCN layers with corresponding activation functions, average pooling, and connect to a fully connected layer to obtain the output. We use the training set to train the model for extracting features and the validation set to evaluate the hyper-parameters of the feature extraction model. Then we use the tuning set to calculate the weights of each modal feature during the fusion stage in the multimodal fused deep learning (MMFDL) models. We compute the weights of the different models after splicing the three learned multi-modal feature outputs using five methods based on machine learning methods and numeric method for combination, which are referred to as Tri_LASSO, Tri_Elastic, Tri_RF, Tri_GB, and Tri_SGD by using LASSO, Elastic Net, Random Forest, Gradient Boosting, and Stochastic Gradient Descent. After obtaining the weights from the layer of weight assignment, all the weighted summation features are then fed into a single concatenated layer to generate the predicted values for the test set. Finally, on the test set, we predict the properties for different tasks using learned features and the assigned weights. To further evaluate the stability and reliability of the results, the model was repeated 15 times using random seed to calculate the Pearson coefficient distributions.

### 2.3.1. Transformer-encoder

Each Transformer encoder layer [70] is composed of a multi-head self-attention sub-layer followed by a position-wise feedforward sub-layer. For the molecule representation $X$, after token embedding and position encoding, it is passed through the encoder to extract features for molecular property prediction. First, let input $X$ for token embedding and obtain the embedding vector $P$. The formula for sinusoidal positional embeddings $PE$ is as follows [70].

$$PE_{(pos,2i)} = \sin\left(\frac{pos}{10000^{\frac{2i}{d}}}\right) \tag{1}$$

$$PE_{(pos,2i+1)} = \cos\left(\frac{pos}{10000^{\frac{2i}{d}}}\right) \tag{2}$$

where $pos$ is the position and $i$ is in the range of $\left[0, \frac{d}{2}\right]$, $d$ is the input dimension. For any fixed offset $k$, $PE_{pos+k}$ can be represented as a linear function of $PE_{pos}$.

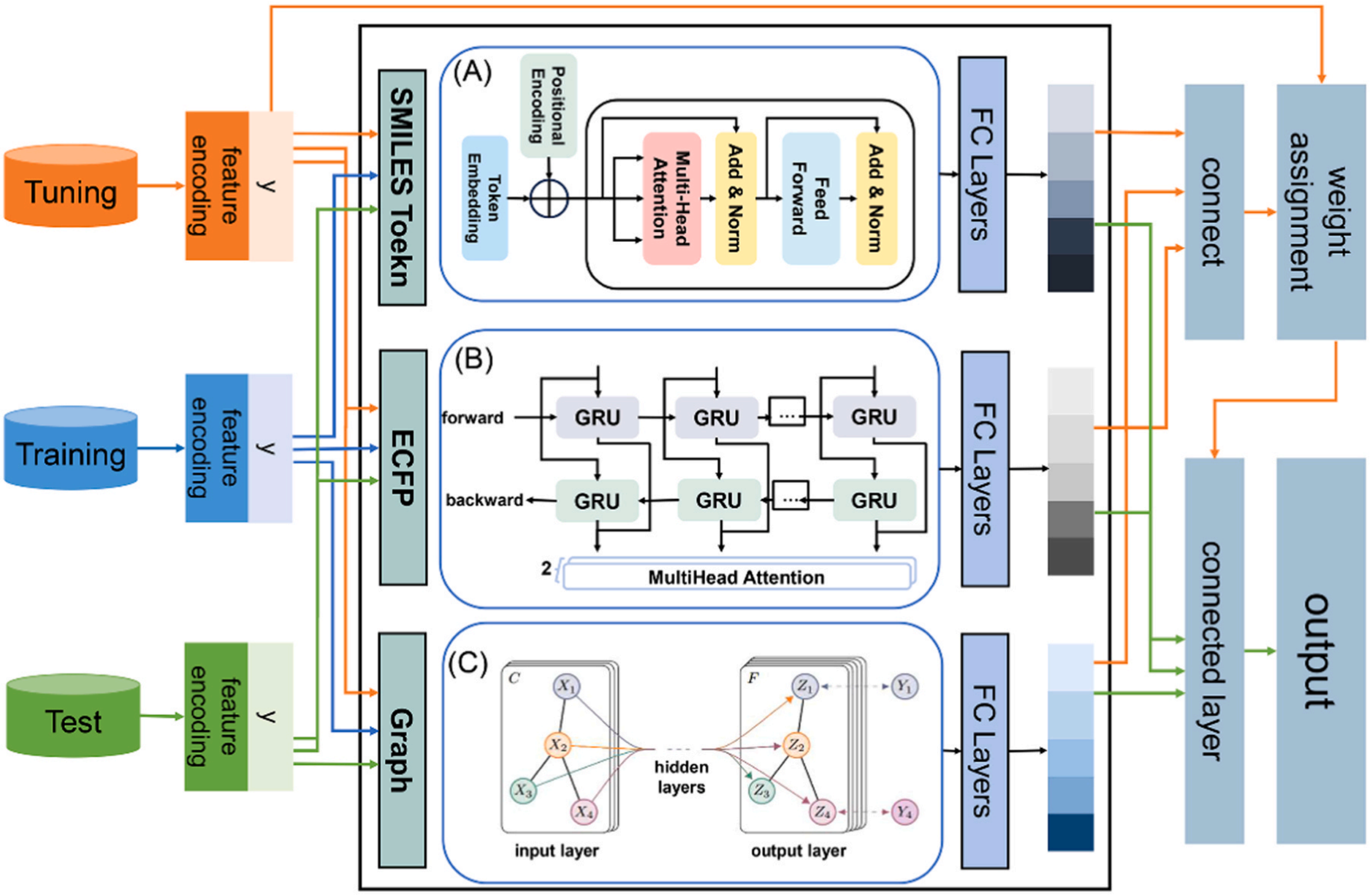

**Fig. 2.** The overall structure of the multimodal fused deep learning (MMFDL) model. SMILES-encoded vectors are processed by (A) Transformer-Encoder, ECFP is processed by (B) BiGRU with MultiHead Attention, and Graph is processed by (C) GCN model. Blue, orange and green color lines represent the data flow in the model for training, tuning, and test sets. The training set is used to train the feature extraction model; the validation set is used to validate the hyperparameters. The tuning set is used to assign the weights for each modal input. The test set is used to validate the prediction performance.

The transformer encoder takes in a matrix $H \in R^{l\times d}$, where $l$ is the molecular sequence length, $d$ is the input dimension. Then three learnable matrix $W_q$,$W_k$,$W_v$ are used to project $H$ into different spaces. Usually, the matrix size of the three matrices is all $R^{d\times d_k}$, where $d_k$ is a hyperparameter. After that, the scaled dot product attention can be calculated by the following equations [71].

$$Q, K, V = HW_q, HW_k, HW_v \tag{3}$$

$$\text{Attention}(Q, K, V) = \text{softmax}\left(\frac{QK^T}{\sqrt{d_k}}\right)V \tag{4}$$

We use several groups of $W_q^{(h)}$,$W_k^{(h)}$,$W_v^{(h)}$ to enhance the ability of self-attention. When several groups are used, it is called multi-head self-attention, the calculation can be formulated as follows [71].

$$Q^{(h)}, K^{(h)}, V^{(h)} = HW_q^{(h)}, HW_k^h, HW_v^h \tag{5}$$

$$head^{(h)} = Q^{(h)}, K^{(h)}, V^{(h)} \tag{6}$$

$$\text{MultiHead}(H) = \left[head^{(1)}...head^{(n)}\right]W_O \tag{7}$$

where $n$ is the number of heads, and the superscript $h$ represents the head index. $[head^{(1)}...head^{(n)}]$ means concatenation in the last dimension. Usually $d_k \times n = k$, which means the output of $[head^{(1)}...head^{(n)}]$ will be of size $R^{l\times d}$. $W_O$ is a learnable parameter, which is of size $R^{d\times d}$.

### 2.3.2. BiGRU

We use BiGRU to capture substructure information of molecules. The main idea of GRU is to make every recurrent unit adaptively capture dependencies of different time scales. However, GRU does not well obtain the acquisition of potential relationships between the current character in the ECFP fingerprints and its surrounding chemical information. We adopt a BiGRU, which uses two sublayers to compute forward and backward hidden sequences $h_t^{forward}$ and $h_t^{back}$ respectively. Then, the formula of BiGRU is described as follows [72].

$$h_t^{forward} = GRU\left(x_t, h_{t-1}^{forward}\right) \tag{8}$$

$$h_t^{back} = GRU\left(x_t, h_{t-1}^{back}\right) \tag{9}$$

$$h_t = W^T h_t^{forward} + W^V h_t^{back} \tag{10}$$

$$o_t = \Phi(W^o h_t) \tag{11}$$

where $W^T$ and $W^V$represent the weight coefficients corresponding to the forward hidden state $h_t^{forward}$ and the reverse hidden state $h_t^{back}$ in the bidirectional GRU, respectively; $W^o$ is the weight coefficient between the hidden and output layers.

### 2.3.3. GCN

Formally, $G = (V, E)$ represents the graph of the molecule, where $V$ is a set of n atomic nodes, and each node is represented by a 78-feature vector of atomic properties. $E$ is the set of edges represented by the adjacency matrix of the molecular graph. The multi-layer graph convolution network (GCN) takes the node feature matrix $X$ and the adjacency matrix $A$ as input, and then produces the node-level output. For stability, the normalized propagation rule is defined as follows [31].

$$H^{(l+1)} = \sigma\left(\widetilde{D}^{-\frac{1}{2}}\widetilde{A}\widetilde{D}^{-\frac{1}{2}}H^{(l)}W^{(l)}\right) \tag{12}$$

where, $\widetilde{A} = A + I_N$ is an adjacency matrix of the undirected graph with added self-connections. $\widetilde{D}_{ii} = \sum_i \widetilde{A}_{ii}$is the diagonal node degree matrix. $H^{(l)}$ and $W^{(l)}$ are the learnable parameters of GCN and the output of the first layer respectively. $\sigma(.)$is an activation function.

A layer-wise convolution operation can be approximated as follows [31].

$$Z = \widetilde{D}^{-\frac{1}{2}}\widetilde{A}\widetilde{D}^{-\frac{1}{2}}X\Theta \tag{13}$$

where $\Theta \in R^{C\times F}$ (F is the number of filters of feature maps) is the matrix of filter parameters.

### 2.3.4. Fusion approach

Multimodal models not only require the construction of multimodal data, but the fusion of different modalities is also crucial. Multimodal fusion is the process of filtering, extracting, and combining required features from various sources of data. Our model uses late fusion (Fig. 2), which first processes each modality individually and then integrates their features at a higher level. Since each modality can learn its representation through independent processing, the model can better capture the specific features and relationships of each modality.

We use four machine learning methods and a gradient descent numerical algorithm to calculate the weights of the multi-modal model. The importance calculation of the triple-modal model is calculated as follows.

$$O = Concat\left(O_{Transformer-Encoder}, O_{BiGRU}, O_{GCN}\right) \tag{14}$$

$$W_1, W_2, W_3 = Weight(O) \tag{15}$$

where $O_{Transformer-Encoder}$, $O_{BiGRU}$, and $O_{GCN}$ represent the outputs of the three feature extraction models, respectively, and $O$ represent the tensor of fused molecular features. $W_1$,$W_2$ and $W_3$ represent the weights of the multi-modal model, and $Weight(.)$ as a method of calculating the importance of the triple-modal model.

In the testing phase, each modal weight is multiplied with the newly calculated multimodal features to generate the final prediction value.

$$output = W_1 O'_{Transformer-Encoder} + W_2 O'_{BiGRU} + W_3 O'_{GCN} \tag{16}$$

where $O'_{Transformer-Encoder}$, $O'_{BiGRU}$ and $O'_{GCN}$ represent the outputs of the three feature extraction models in the testing phase, respectively.

## 3. Results and discussion

### 3.1. Performance of single and multi-modal models

To verify the effectiveness of the multimodal model, we constructed both mono-modal models and multimodal models. We proposed to employ SMILES-encoded vectors, ECFP, and molecular graphs to provide complementary representations for the multimodal fused deep learning model. To validate this proposal, we took Delaney and SAMPL training dataset as an example and conducted the uniform manifold approximation and projection (UMAP) dimensionality reduction on the vectors of the FC layers (referred to Fig. 2) of each modal. As shown in Fig. 3, the three modal distribution shows both overlapping and non-overlapping regions for the vectors of hidden FC layers, indicating the presence of the shared as well as independently complementary information across three modalities.

The performance of the models was compared by calculating the root mean square error (RMSE), mean absolute error (MAE), and Pearson coefficient on the test set. As observed from the RMSE values in Table 1 and MAE values in Table 2, Tri_LASSO, Tri_Elastic, Tri_RF, and Tri_SGD show lower RMSE and MAE values than the best individual model in the SAMPL and DataWarrior datasets. The Tri_SGD model achieves the lowest RMSE and MAE values among all tasks. Tri_SGD is a robust fusion method that can effectively handle the characteristics of different data sets, displaying better results than the other four fusion methods. Multimodal learning can achieve better prediction than mono-modal learning.

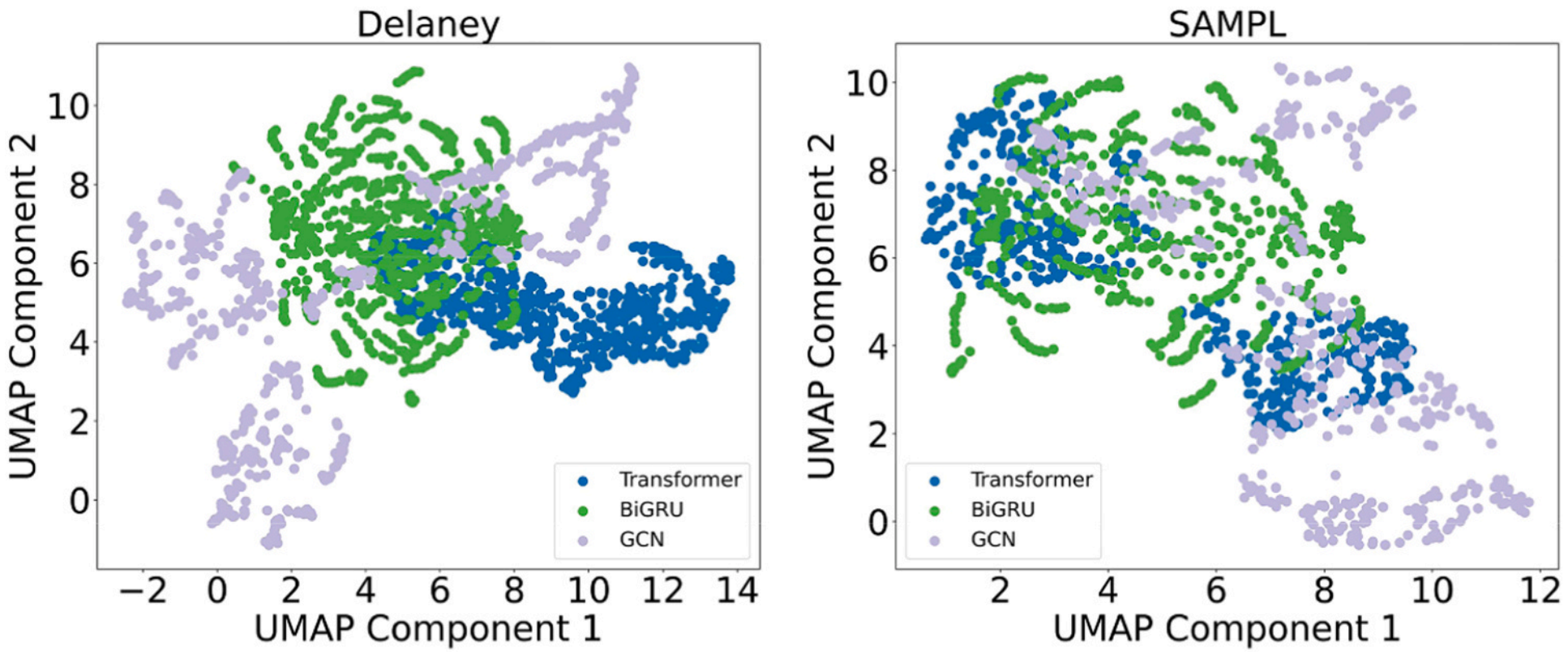

**Fig. 3.** UMAP dimensionality reduction diagram of the vectors of hidden FC layers for the three modalities based on training set of Delaney and SAMPL datasets.

**Table 1**
RMSE performance of different data sets on 8 different models.

| Dataset | Delaney | Llinas2020 | Lipophilicity | SAMPL | BACE | DataWarrior |
|---|---|---|---|---|---|---|
| Transformer | 0.671 | 1.208 | 0.937 | 1.407 | 1.177 | 2.672 |
| BiGRU | 1.259 | 1.339 | 0.863 | 1.788 | 0.806 | 2.555 |
| GCN | 0.858 | 1.337 | 0.911 | 2.429 | 1.075 | 3.195 |
| Tri_LASSO | 0.811 | 1.193 | 0.782 | 1.099 | 0.981 | 2.337 |
| Tri_Elastic | 0.872 | 1.308 | 0.949 | **1.103** | 1.669 | 2.309 |
| Tri_RF | 0.804 | 1.173 | 0.848 | 1.341 | 0.978 | 2.215 |
| Tri_GB | 0.759 | 1.145 | 0.827 | 1.359 | 0.982 | 2.209 |
| Tri_SGD | **0.620** | **1.065** | **0.725** | **1.103** | **0.762** | **2.164** |

**Table 2**
MAE performance of different data sets on 8 different models.

| Dataset | Delaney | Llinas2020 | Lipophilicity | SAMPL | BACE | DataWarrior |
|---|---|---|---|---|---|---|
| Transformer | 0.489 | 0.958 | 0.737 | 1.078 | 0.936 | 1.920 |
| BiGRU | 0.932 | 1.092 | 0.630 | 1.258 | 0.552 | 1.758 |
| GCN | 0.675 | 1.095 | 0.737 | 1.803 | 0.878 | 2.441 |
| Tri_LASSO | 0.568 | 0.955 | 0.616 | 0.855 | 0.770 | 1.706 |
| Tri_Elastic | 0.616 | 1.068 | 0.788 | 0.848 | 1.491 | 1.673 |
| Tri_RF | 0.613 | 0.933 | 0.639 | 1.00 | 0.723 | 1.562 |
| Tri_GB | 0.577 | 0.912 | 0.626 | 1.013 | 0.726 | 1.552 |
| Tri_SGD | **0.470** | **0.844** | **0.565** | **0.822** | **0.530** | **1.540** |

The Pearson coefficient was calculated to evaluate the model performance on the test sets. As shown in Table 3, it can be observed that Tri_LASSO, Tri_Elastic, Tri_RF, Tri_GB, and Tri_SGD show consistent performance on the Lipophilicity, SAMPL, BACE, and pKa DataWarrior datasets. The Tri_SGD exhibits a notable improvement in the Llinas2020 and Lipophilicity datasets. Among mono-modal learning, only Transformer can obtain comparable accuracy in Delaney and SAMPL datasets. Herein, we demonstrate the application of multimodal deep learning. More accurate prediction of pKa has been reported by Mansouri et al. [73]. We conducted k-fold cross validation using Delaney dataset. The cross validation results shown in Fig. S2 indicated that multimodal fusion method performed well in reliability in the 5-fold cross validation. Overall, Tri_SGD achieves optimal values for RMSE, MAE, and Pearson coefficients across all tasks.

To measure the similarity between the true and predicted value, we calculated the cosine similarity between the true value and the predicted value of each model. As shown in Fig. S3, across different tasks, the cosine similarity of the true values and predicted values of Tri_SGD is higher than that of any single modal model, suggesting the predicted values of Tri_SGD are generally close to the true values. Tri_RF achieved the highest value in the SAMPL dataset, and Tri_SGD is very close to Tri_RF. The cosine similarity of Tri_SGD reaches the highest value for five datasets among six datasets.

**Table 3**
Pearson coefficients of different data sets on 8 different models.

| Dataset | Delaney | Llinas2020 | Lipophilicity | SAMPL | BACE | DataWarrior |
|---|---|---|---|---|---|---|
| Transformer | 0.95 | 0.56 | 0.65 | 0.93 | 0.70 | 0.65 |
| BiGRU | 0.80 | 0.55 | 0.71 | 0.87 | 0.80 | 0.68 |
| GCN | 0.92 | 0.69 | 0.64 | 0.74 | 0.59 | 0.38 |
| Tri_LASSO | 0.94 | 0.67 | 0.75 | **0.95** | 0.75 | 0.75 |
| Tri_Elastic | 0.95 | 0.66 | 0.76 | **0.95** | 0.76 | 0.75 |
| Tri_RF | 0.93 | 0.66 | 0.72 | **0.95** | 0.75 | 0.75 |
| Tri_GB | 0.94 | 0.67 | 0.73 | **0.95** | 0.75 | 0.75 |
| Tri_SGD | **0.96** | **0.73** | **0.79** | **0.95** | **0.82** | **0.76** |

### 3.2. Model reliability

To further evaluate the stability and reliability of the results, the model was repeated 15 times to calculate the Pearson coefficient distributions of the models. Multiple calculations of the Pearson coefficients provide a better reflection of the overall stability of the model's predictions. As shown in Fig. 4, the performance of different mono-modal models shows a relatively wider distribution in the single-molecule datasets. In addition, the performance of mono-modal models varies across different datasets and does not achieve reliable results as the multimodal learning models. The multimodal fusion method Tri_SGD proposed in this study has the best Pearson coefficient across different datasets, except the SAMPL dataset, where Tri_Elastic has the best Pearson coefficient value.

Overall, through 15 rounds of repeated calculations, the multimodal fusion method performs outstandingly in reliability. Especially, Tri_SGD can significantly enhance the overall stability and reliability of the model, has better generalization ability, and is more adaptable to different data features.

As shown in Fig. 4 and Fig. S4, it can be seen that the mono-modal models show the relatively lower Pearson coefficients, and larger RMSE and MAE, than that of multimodal fusion methods. The proposed fusion method Tr_SGD has the best Pearson coefficient, RMSE and MAE across different datasets, except the SAMPL dataset, where Tri_Elastic has the best Pearson coefficient value and Tri_LASSO has the smallest RMSE and MAE values. Overall, multimodal fusion methods perform well in terms of reliability.

### 3.3. Analysis using feature distribution

To observe the relationship between the input features, we visualized the spatial distribution of SMILES-encoded vectors and ECFP fingerprints by using the dimensionality reduction method of uniform manifold approximation and projection (UMAP). Graph data is not suitable for dimensionality reduction because of its combined nodes and edges data format. The unmatched distribution between training and test sets explained the poor performance of mono-modal learning models. As shown in Fig. 5A, we find that SMILES-encoded vectors show well-overlapping spaces between the test set and the training set for four datasets other than the Llinas2020 and BACE datasets. Consistent with the RMSE results, Transformer does not obtain satisfactory RMSE and MAE on Llinas2020 and BACE dataset because the deep learning methods are generally better at interpolation than extrapolation. In Fig. 5B, the ECFP fingerprints do not overlap well between the training and test sets in SAMPL. Therefore, the RMSE of BiGRU is higher than that of Transformer in SAMPL.

From UMAP, we can notice that the single molecular representation can represent parts of information. In some cases, the inability to achieve a consistent UMAP embedding space between the training and test sets has led to poorer predictive results on the test set. From molecular feature distribution, we can infer that the single modal learning models will tend to give poor performance if the test set does not well-overlap with the training set. In contrast, multi-modal learning still provides better performance in both Llinas2020 and BACE datasets than any mono-modal learning models.

To further evaluate the impact of input features on model performance, we calculated the maximum similarity and distance of k-NN for six single-molecule datasets. As can be seen from Fig. 6, when the distribution overlap of the test set and the training set is low, its maximum similarity is also low. For example, the Llinas2020 dataset has poor spatial overlap between SMILES-encoded vectors and ECFP fingerprints in UMAP, and also has a lower similarity distribution, leading Transformer and BiGRU to produce the lowest Pearson coefficients. In contrast, for the SAMPL dataset, the maximum similarity of the SMILES-encoded vectors is better than ECFP fingerprints. Therefore, the RMSE and MAE of the mono-modal Transformer are lower than BiGRU, and the Pearson coefficient is higher than BiGRU. The same conclusion also can be obtained from the k-NN distance distribution. We computed the average distance between one molecule in test set and five molecules in the training set with the shortest distance. SAMPL shows the shortest distance in the SMILES-encoded vectors, and obtains the best prediction among three mono-modal learning method.

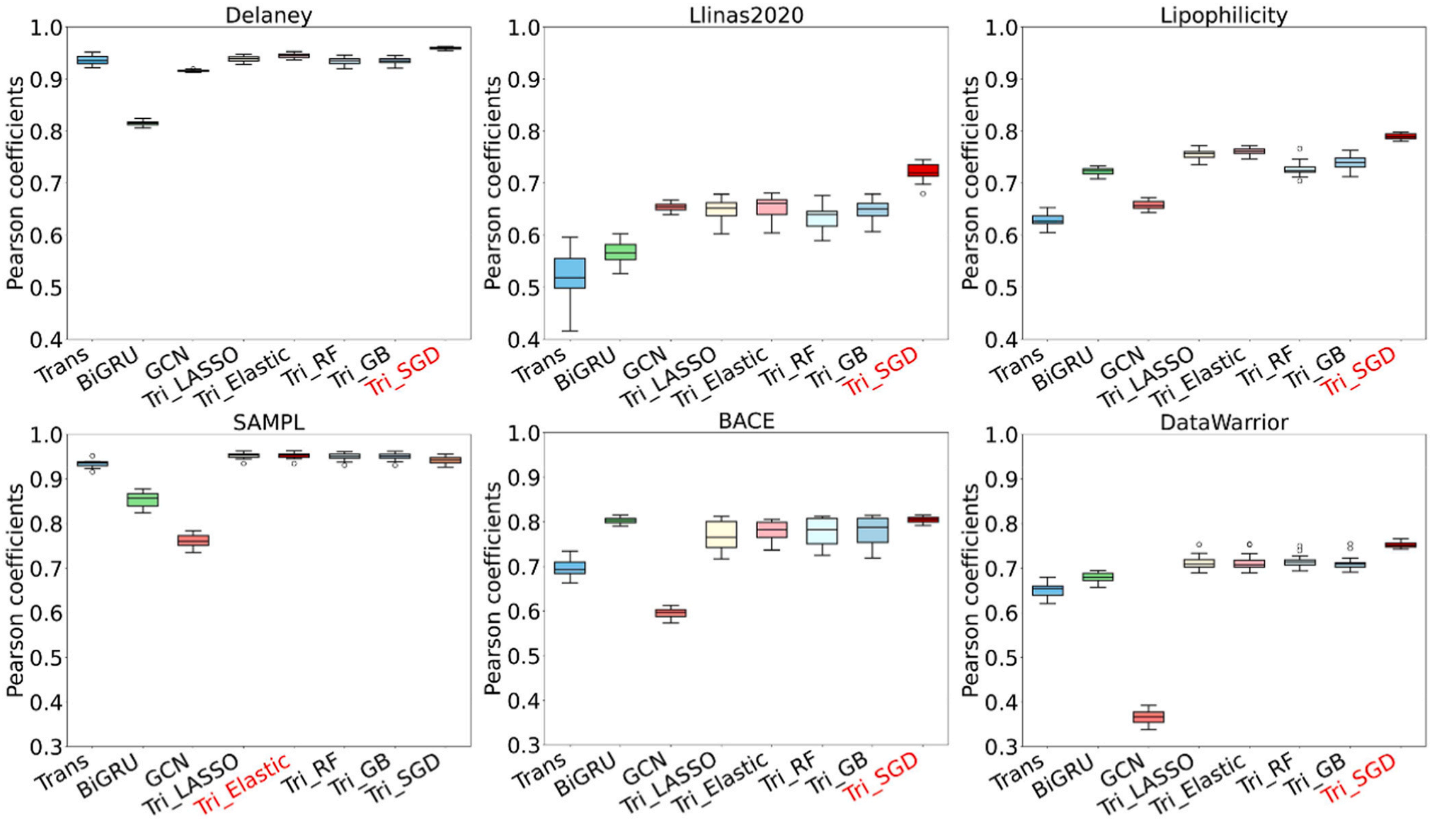

**Fig. 4.** Pearson coefficients of different datasets using three mono-modal learning methods and five fused triple-modal learning methods.

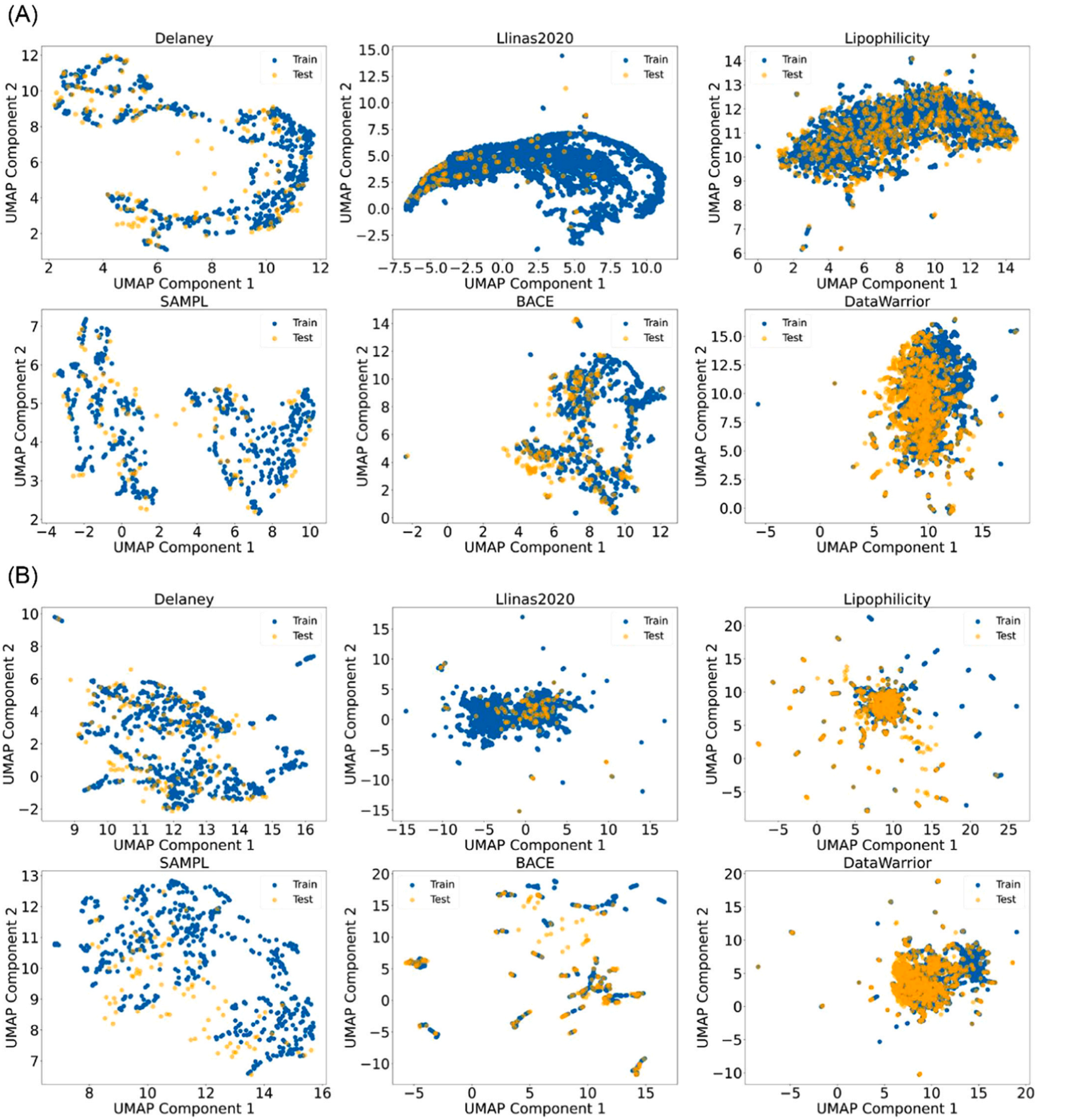

**Fig. 5.** UMAP diagram of SMILES-encoded vectors and ECFP. (A) UMAP diagram of SMILES-encoded vectors in the training and testing sets. (B) UMAP diagram of ECFP in training and testing set.The training and test dataset is colored in blue and orange.

By combining results from Fig. 4, Fig. 5, and Fig. 6, we can observe that when one molecular representation method leads to barely overlapping regions between the training and test sets, the other representation method may result in overlap between the training and test sets. This indicates two points. Firstly, when the training and test sets overlap in molecular representation space, it enables the prediction model to better utilize training data, leading to improved prediction performance. Secondly, different molecular representation methods express different chemical spaces, necessitating a multimodal model to integrate multiple representation methods. The integrating multiple modalities to capture a more comprehensive and representative view of chemical space will lead to improved performance. Therefore, we propose the multimodal fusion methods to fuse different features to improve the performance in drug property prediction.

### 3.4. *Weights analysis in fusion*

The stability and reliability of multimodal fused deep learning

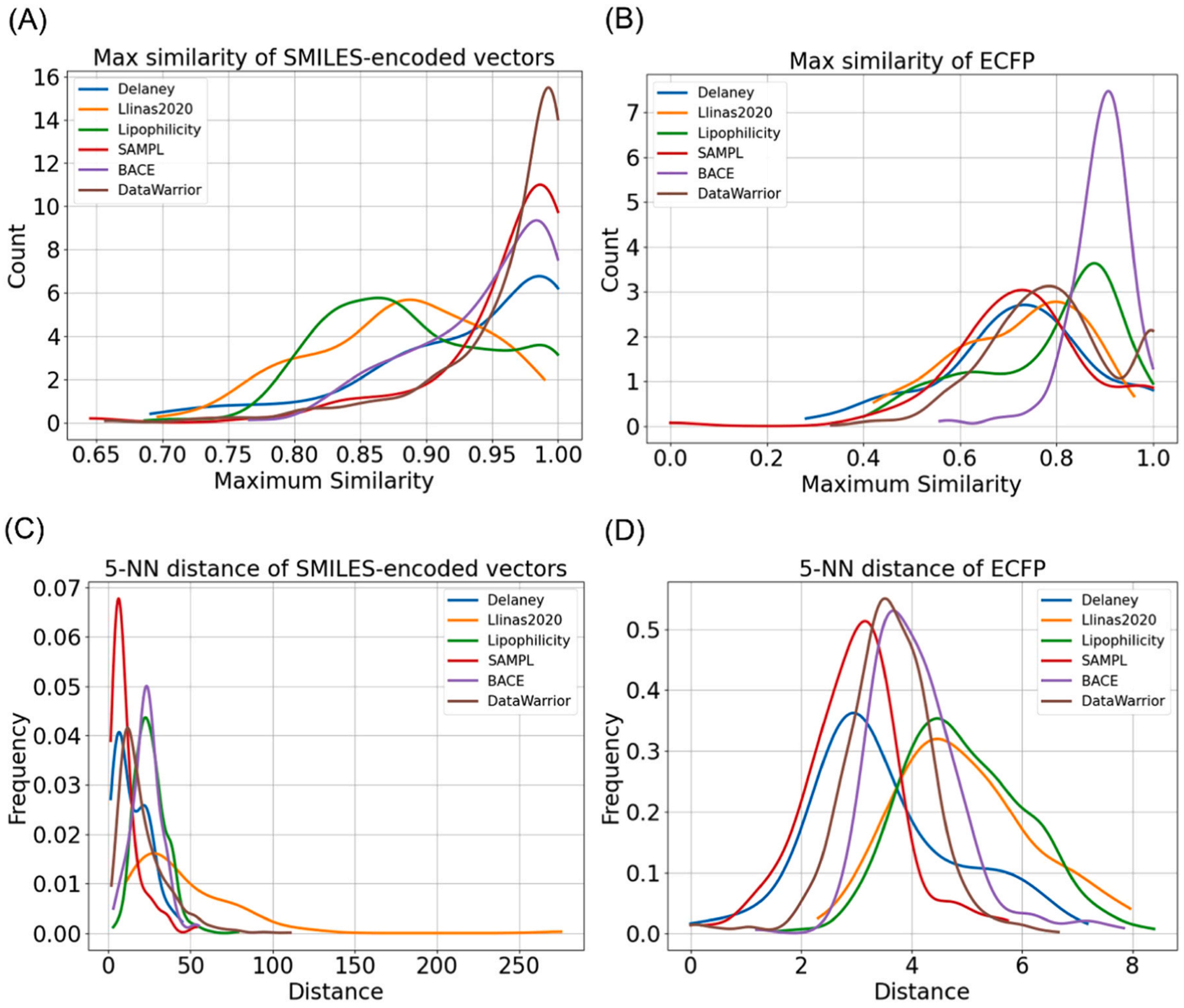

**Fig. 6.** Distribution diagram of maximum similarity and 5-NN distance between SMILES-encoded vectors and ECFP fingerprints. (A) The maximum similarity distribution of the training set and test set of SMILES-encoded vectors. (B) The maximum similarity distribution of the training set and test set of ECFP fingerprints. (C) The 5-NN distance distribution of the training set and test set of SMILES-encoded vectors. (D) The 5-NN distance distribution of the training set and test set of ECFP fingerprints.

(MMFDL) models are better than mono-modal models. We further explore the weight differences among the five fusion methods (Tri_LASSO, Tri_Elastic, Tri_RF, Tri_GB, and Tri_SGD) from a weight distribution perspective. In multimodal learning, some modalities may provide better predictions than others. As shown in Fig. 7, Tri_LASSO, Tri_Elastic, Tri_RF and Tri_GB exhibit unbalanced weight distribution, especially in terms of the weight of graph information. This unbalanced weight distribution may lead to insufficient consideration of graph features, affecting the overall performance and generalization ability of the model. Tri_LASSO and Tri_Elastic give higher weight to the Transformer model in the Delaney, SAMPL and DataWarrior datasets, while Tri_RF and Tri_GB give higher weight to the BiGRU model in the Llinas2020 and Lipophilicity datasets. Overall, Tri_SGD is able to achieve relatively balanced weights in datasets of Delaney, Llinas2020, Lipophilicity, and BACE. According to Fig. 4, and Fig. S3, it can be seen that Tri_SGD basically contains the best values for all tasks, except DataWarrior. Tri_SGD fusion method makes full use of the three types of features, indicating that an appropriate allocation strategy can improve the accuracy and generalization ability of the model. The average aggregation over three input features is also computed for comparison using Delaney and BACE dataset. As shown in Fig. S5, the average aggregation method produces the larger RMSE and lower Pearson coefficient than our proposed fusion method. The results highlight the importance of selecting appropriate methods based on the specific characteristics and needs of each application domain.

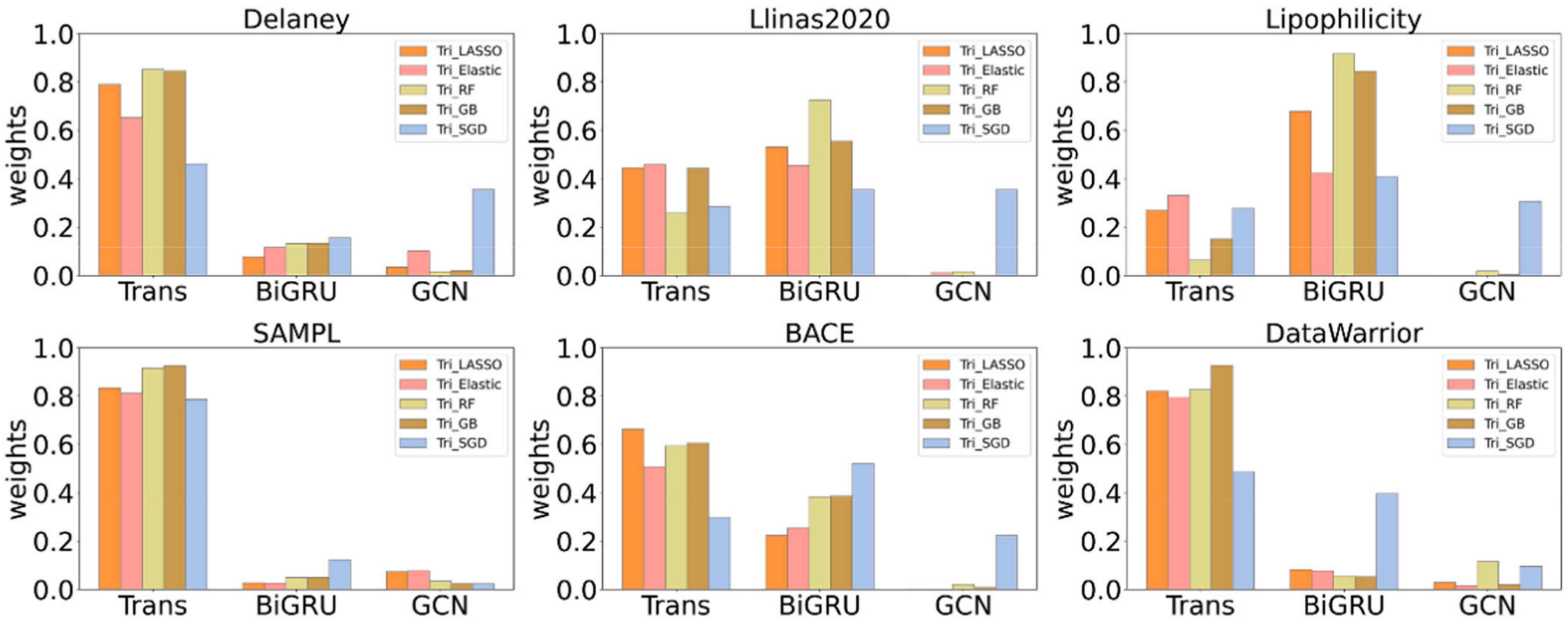

**Fig. 7.** Weights distribution for each modal input of triple-modal learning methods in different datasets.

SGD is well-suited for online learning because it can handle streaming data in real-time and perform continuous updates, while machine learning methods are more suitable for post-processing or offline learning tasks. The fusion methods Tri_LASSO, Tri_Elastic, Tri_RF and Tri_GB allocate more weights to modalities that can process sequence data and assign low or 0 weights to GCN, while Tri_SGD is able to comprehensively integrate modal information. Tri_LASSO and Tri_Elastic are linear regression methods that use regularization. In the presence of weakly correlated features in the data, these regularization terms may lead feature weights to be zero. SMILES encoding vectors and ECFP fingerprints are one dimensional data with a certain correlation, so that these two types of features are given higher weight. Tri_RF and Tri_GB are nonlinear methods that can handle a certain degree of irrelevant information and tolerate a certain amount of sparse data. When feature correlation is poor, the usefulness of tree-based models is also limited. Comparing the results of five methods, we find that ignoring one modality may lead to poorer performance. Tri_SGD is an optimization algorithm that can gradually adjust parameters to adapt to data characteristics during the on-the-fly optimization process. This adaptive function can effectively capture the relationship of triple inputs. Among the five fusion methods, only Tri_SGD can assign a more appropriate weight distribution and exhibit the best performance in each dataset. The results remind us that an effective fusion method should fully exploit the complementary information of each modality input.

### 3.5. Generalization ability validation

#### 3.5.1. Generalization ability validation using protein-ligand complex molecules

Our proposed multimodal fusion method achieved the best performance in single-molecule datasets of solubility, lipophilicity, inhibitory activity, and acidity/basicity. To verify the applicability of this method, we selected the refined set of the PDBbind v2020 dataset for verification. The PDBbind v2020 dataset has a total of 23496 binding constants of biomolecular complexes, including protein ligands, nucleic acid ligands, protein-nucleic acids, proteins, and protein complexes. We use the PDBbind2020 refined dataset to evaluate the generalization ability of MMFDL. The PDBbind2020 refined dataset screens better quality protein-ligand complexes from the general dataset, containing a total of 5316 selected protein-small molecule complexes from which binding constants were obtained. We transformed PDB files into three molecular representations. After transformation, 5089 molecules can be successfully converted into three specified descriptors, and others are removed due to the existence of unrecognized molecules in the RDKit. As shown in Table 4, Tri_Elastic shows the minimum RMSE and MAE values, and the results of Tri_SGD are very close to the best results of Tri_Elastic.

**Table 4**
Performance of RMSE, MAE, and Pearson on PDBbind2020 refined dataset.

| Dataset | RMSE | MAE | Pearson coefficient |
|---|---|---|---|
| Transformer | 1.807 | 1.420 | 0.56 |
| BiGRU | 1.385 | 1.076 | 0.72 |
| GCN | 1.986 | 1.570 | 0.64 |
| Tri_LASSO | 1.536 | 1.200 | 0.67 |
| Tri_Elastic | **1.349** | **1.048** | 0.72 |
| Tri_RF | 1.408 | 1.108 | 0.72 |
| Tri_GB | 1.400 | 1.101 | **0.73** |
| Tri_SGD | 1.358 | 1.067 | **0.73** |

Both Tri_GB and Tri_SGD achieve the best Pearson coefficient, but Tri_SGD provides relatively smaller RMSE and MAE. By leveraging multiple modalities, multi-modal learning enables the model to learn more comprehensive and robust representations of the biological data, leading to better generalization to unseen examples.

To verify the stability of the multimodal model in the PDBbind2020 refined dataset, we also conducted 15 rounds of random seed testing. As observed in Fig. 8 A and Fig. 8B, the Pearson coefficient of single modal models in this dataset fluctuates, while the proposed multimodal fusion deep learning model shows relatively stable performance. The proposed multimodal fusion deep learning can be extended to predict the binding affinity for the protein-ligand complex. Using the current hyperparameters, Tri_RF produces the highest average value of Pearson coefficient (Fig. 8 A), and Tri_RF, Tri_GB and Tri_SGD can provide the highest median values of Pearson coefficient (Fig. 8B). The result displays the possible applications in other biological problems. Combining the evaluation of RMSE, MAE, and Pearson coefficient, the Tri_SGD method demonstrated good accuracy and robustness on the PDBbind2020 refined dataset, and can better predict protein-ligand binding affinity. We did not optimize the hyperparameters or molecular descriptions when applying our model to the PDBbind dataset and just extended the input dimension from the single-molecule dataset to protein-ligand complex molecules. In this study, we put our emphasis on the multi-modal learning methods and hence more accurate results would be reached after optimizing the models and selecting more proper inputs.

#### 3.5.2. Generalization ability validation using 3D molecular descriptors

The proposed multimodal fusion deep learning (MMFDL) model is not restricted to processing only the three mentioned molecular representations (SMILES, ECFP, and molecular graphs). This framework can be extended to incorporate other types of molecular representations, such as 3D molecular structures, computer learned molecular descriptors, or any other relevant data modalities.

To demonstrate the generalization ability in incorporating other types of molecular representations, we replaced one of the original molecular representations, ECFP fingerprints, with a 3D molecular representation. E3FP fingerprint was employed alongside SMILES-encoded vectors and molecular graph in the BACE dataset. E3FP is one 3D molecular fingerprinting method based on the three-dimensional structural information of molecules, which takes into account the spatial arrangement and stereochemical characteristics of molecules. Results in Fig. 8 C indicated that E3FP produced lower RMSE for two multimodal learning methods (Tri_LASSO and Tri_Elastic), while ECFP yielded better results for others. OnionNet network features were compared with ECFP in the PDBbind2020 refined dataset. OnionNet is a convolutional neural network based on multi-layer intermolecular contacts. The input features are the atom-pair contacts in different ranges of distances based on element-pair-specific contacts between ligand and protein atoms. ECFP outperforms OnionNet features across the proposed methods. The incorporation of 3D information did not lead to significant performance improvements, potentially due to the suboptimal choice of the BiGRU model in capturing complex features of 3D molecular structures. Future research is suggested to explore optimization of the multimodal fusion deep learning model architecture. These findings contribute to the understanding of molecular representation and multimodal learning methods in drug discovery.

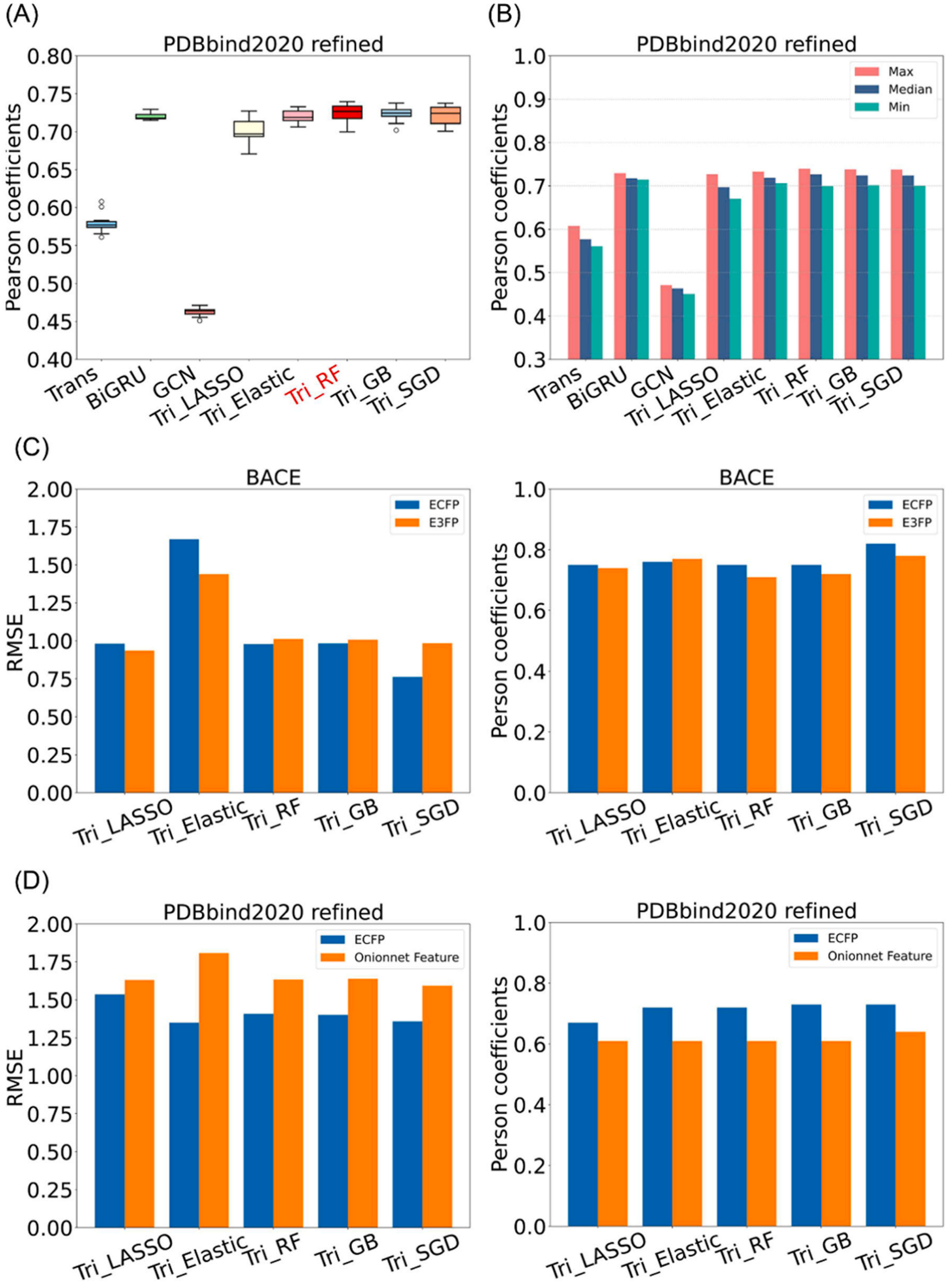

**Fig. 8.** RMSE and Pearson coefficients for generalization ability validation. (A) The box plot of the repeatedly calculated Pearson coefficient, and (B) the histogram recording the maximum, median, and minimum values of the Pearson coefficient. (C) RMSE and Person coefficients of the BACE dataset using ECFP and E3FP. (D) RMSE and Pearson coefficients of PDBbind2020 refined dataset using ECFP and Onionnet feature.

### 3.6. Noise resistance capability of multimodal model

To verify the noise immunity of the model, we did a test on the SAMPL dataset by adding different noise levels of 0.05, 0.1, 0.2, and 0.5 to the input features of all the models. The error bar was computed from the 15 times repeated training. SMILES-encoded vectors were added noise with random values from a dictionary size of 1 to 34, and ECFP fingerprints and graph feature vectors were added noise at random bit positions (0 or 1). In Fig. 9, graph data is susceptible to noise, and it leads single modal model GCN to the highest RMSE and MAE, and lowest Pearson coefficients after adding the noise. Figs. 9A and 9B are histograms of RMSE and MAE values of mono-single modals and fused triple-modal learning methods after adding different proportions of noise to the SAMPL dataset. With the increase in the level of noise, BiGRU produces the largest RMSE and MAE values, followed by GCN. With the increase in the level of noise, the performance of BiGRU also becomes worse. In contrast, the multimodal learning model shows remarkable noise resistance. As shown in Fig. 9C, the Pearson coefficients of all models decrease with the increasing of the ratio of noise. All five fusion methods can provide certain satisfactory results when the ratio of noise is less than ~0.2. From the perspective of RMSE and MAE, the performances of Tri_LASSO, Tri_Elastic, Tri_RF, and Tri_GB are similar at different noise levels, while Tri_SGD has the lowest RMSE and MAE values, indicating the proper fusion method can enhance the capability of noise resistance. The multi-modal fusion deep learning model demonstrates increased robustness in the presence of noisy data, with Tri_SGD maintaining predictive accuracy while being more resilient to noise. The result indicates that it is very essential to have proper fusion methods to keep useful information from each modal input.

Predicting the activity, solubility, toxicity, and other properties of drug molecules, can help researchers design new drug molecules in a targeted manner. We construct a multimodal fusion deep learning (MMFDL) model that can predict molecular properties using complementary information from chemical language and molecular graphs. We use five fusion approaches to evaluate the proposed three-modality model on six single-molecule datasets, including Delaney, Llinas2020, Lipophilicty, SAMPL, BACE, and DataWarrior, and the results show that MMFDL outperforms mono-modal models in accuracy and reliability, with the Tri_SGD fusion method further improving the ability of multi-modal learning models. In addition, we verified the generalization and noise resistance ability of MMFDL model, and MMFDL showed good accuracy and robustness. The feature distribution in chemical space demonstrates the need of appropriate fusion methods based on input features to obtain complementary information.

MMFDL model can be extended to incorporate other types of molecular representations. However, "No one size fits all" highlights the importance of considering the specific characteristics and requirements of each application domain when selecting appropriate modeling approaches. In the context of molecular representation and deep learning, this principle underscores the need for tailored modeling strategies that are well-suited to the particular features and complexities of different molecular datasets and tasks. The findings from our investigation suggest the importance of selecting appropriate deep learning architectures that possess the requisite capacity to effectively process and utilize the information encoded within the molecular representations. For the future applications, we suggest systematically evaluating and selecting appropriate combinations of molecular representations and deep learning models based on their compatibility and effectiveness in addressing specific bioinformatics tasks. Other fusion methods and architectures are expected to utilize different biological information.

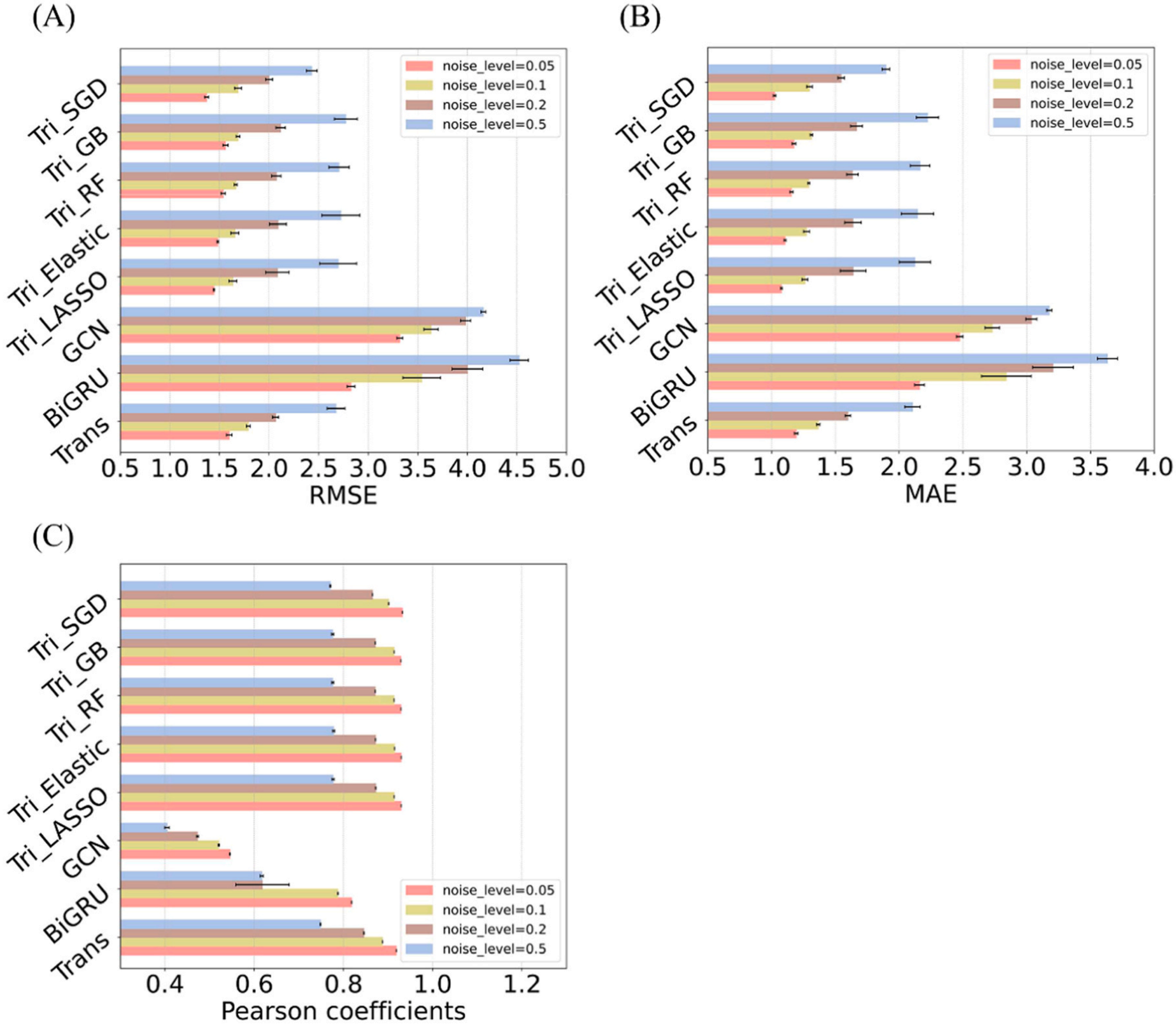

**Fig. 9.** Performance of models in noise resistance test. RMSE, MAE, and Pearson coefficient after adding different ratios of noise on the SAMPL dataset.

## 4. Conclusions

We construct a multimodal fused deep learning (MMFDL) model that can utilize the complementary information of chemical language and molecular graphs to predict molecular properties. In this study, we apply SMILES-encoded vectors, ECFP, and graphs to represent molecules and adopt Transformer-Encoder, BiGRU, and GCN for each modal learning, respectively. We validate the prediction performance of the MMFDL model in comparison with mono-modal by using six single-molecule datasets. The result indicates that the proposed MMFDL model not only improves the prediction accuracy and stability but also increases the ability of noise resistance by properly leveraging different sources of information. By assigning the weights in the fusion of triple modal features in the MMFDL model, the Tri_SGD fusion method further improves the performance of multimodal learning models in dealing with uncorrelated data sources, such as chemical language and molecular graphs. Moreover, we demonstrate the generalization ability in predicting binding affinity for protein-ligand complex. To sum up, we demonstrate that integrating information from different modalities can enhance prediction accuracy, generalization capability, and noise resistance. This highlights the power of leveraging multiple modalities in improving the accuracy and robustness of the predictions. Given the inherent need to handle complex biological big data in the drug discovery process, multimodal learning is expected to have more significant impact on drug discovery.

## Funding

This work was supported by the National Natural Science Foundation of China (22003020, 62373172 and 12074151), the Natural Science Foundation of Jiangsu Province (BK20191032), and Changzhou Sci. & Tech. Program (CJ20200045), Postgraduate Research & Practice Innovation Program of Jiangsu Province (SJCX22_1480).

## CRediT authorship contribution statement

**Xiaohua Lu:** Data curation, Writing – original draft. **Liangxu Xie:** Conceptualization, Writing – review & editing. **Lei Xu:** Formal analysis, Methodology. **Rongzhi Mao:** Investigation. **Xiaojun Xu:** Software, Validation. **Shan Chang:** Funding acquisition, Supervision.

## Declaration of Competing Interest

The authors declare that they have no known competing financial interests

## Data Availability

Data and source codes are available in a GitHub repository and can be accessed via the link: https://github.com/AIMedDrug/MMFDL.git. We have also included Jupyter notebooks in the GitHub repository that can be executed with Google Colab. The protocol and links to access these scripts are provided within the repository.

## Appendix A. Supporting information

Supplementary data associated with this article can be found in the online version at doi:10.1016/j.csbj.2024.04.030.